\documentclass[sigconf, 9pt]{acmart}
\usepackage{booktabs}
\usepackage{multirow}
\usepackage{hyperref}
\usepackage{microtype}
\usepackage{algorithm}
\usepackage{algorithmic}
\usepackage{bbm}
\usepackage{subfig}
\usepackage{hyperref}
\usepackage{graphicx}
\usepackage{amsmath}
\usepackage{threeparttable}

\AtBeginDocument{%
  }

\renewcommand\footnotetextcopyrightpermission[1]{}

\begin{document}

\title{DeepGate3: Towards Scalable Circuit Representation Learning}

\author{Zhengyuan Shi\textsuperscript{1*}, Ziyang Zheng\textsuperscript{1*}, Sadaf Khan\textsuperscript{1}, Jianyuan Zhong\textsuperscript{1}, Min Li\textsuperscript{2}, Qiang Xu\textsuperscript{1}}
\email{{zyshi21, zyzheng23, skhan, jyzhong, qxu}@cse.cuhk.edu.hk}
\affiliation{%
  \institution{\textsuperscript{1}The Chinese University of Hong Kong, \textsuperscript{2}Noah’s Ark Lab, Huawei}
  \country{}
}

\renewcommand{\shortauthors}{Zhengyuan Shi, Ziyang Zheng, Sadaf Khan, Jianyuan Zhong, Min Li, Qiang Xu}

\begin{abstract}

Circuit representation learning has shown promising results in advancing the field of Electronic Design Automation (EDA). Existing models, such as DeepGate Family, primarily utilize Graph Neural Networks (GNNs) to encode circuit netlists into gate-level embeddings. However, the scalability of GNN-based models is fundamentally constrained by architectural limitations, impacting their ability to generalize across diverse and complex circuit designs. To address these challenges, we introduce DeepGate3, an enhanced architecture that integrates Transformer modules following the initial GNN processing. This novel architecture not only retains the robust gate-level representation capabilities of its predecessor, DeepGate2, but also enhances them with the ability to model subcircuits through a novel pooling transformer mechanism. DeepGate3 is further refined with multiple innovative supervision tasks, significantly enhancing its learning process and enabling superior representation of both gate-level and subcircuit structures. Our experiments demonstrate marked improvements in scalability and generalizability over traditional GNN-based approaches, establishing a significant step forward in circuit representation learning technology. 

\end{abstract}

\maketitle
\let\thefootnote\relax\footnotetext{* Both authors contributed equally to this research.}

\section{Introduction}

Pre-training scalable representation learning models and subsequently fine-tuning them for diverse downstream tasks has emerged as a transformative paradigm in artificial intelligence (AI). This evolution is facilitated by the expansion of training data~\cite{chowdhery2023palm}, allowing these models not only to enhance performance on existing tasks but also to demonstrate exceptional capabilities in novel applications. For instance, large language models like GPT-4\cite{achiam2023gpt}, T5\cite{t5} and Roberta\cite{liu2019roberta} demonstrate near-human proficiency in text comprehension and generation, inspiring similar methodologies in other domains such as computer vision \cite{wang2023internimage} and graph analysis \cite{liu2023towards}.


In the realm of Electronic Design Automation (EDA), the quest for effective circuit representation learning has also garnered significant attention. Established approaches such as the DeepGate family \cite{li2022deepgate, shi2023deepgate2, khan2023deepseq}, Gamora \cite{wu2023gamora}, and HOGA \cite{deng2024less} employ Graph Neural Networks (GNNs) to translate circuit netlists into gate-level embeddings, successfully supporting a variety of downstream EDA tasks. Despite their notable successes, the scalability of GNN-based models is fundamentally constrained by architectural limitations. 

That is, GNN-based models, while adept at handling structured data, fail to align with traditional scaling laws \cite{kaplan2020scaling}. Merely increasing the size of training datasets does not proportionately enhance performance, a limitation substantiated by recent findings \cite{liu2024neural}. Moreover, the discriminative capacity of GNNs is often insufficient; their inherent message-passing mechanism can result in information distortion as messages traverse long paths within large graphs, leading to challenges in distinguishing between similar graph structures \cite{alon2020bottleneck, topping2021understanding}. This limitation is highlighted by the difficulties faced by common GNN architectures such as GCN \cite{kipf2016semi} and GraphSAGE \cite{hamilton2017inductive}, which struggle to differentiate similar graph structures \cite{xu2018powerful}.

This introduces a critical challenge: \emph{How can we scale circuit representation models to effectively leverage large amounts of training data while continuing to improve model performance and generalization capabilities?}


Inspired by the success of Transformer$^{\text{1}}$\footnote{1. Transformer~\cite{vaswani2017attention}, known for its universal approximation capabilities~\cite{yun2019transformers} and remarkable scalability \cite{kaplan2020scaling}, is the \emph{de-facto} neural network architecture for processing a variety of data types, including natural language, speech, and images.}-based graph learning models \cite{zhang2020graph, ying2021transformers, rong2020self, xia2024opengraph}, we propose DeepGate3, an enhancement over the previous circuit representation learning model, DeepGate2, by integrating Transformer blocks with GNN. This integration marries the flexible representation capabilities of GNNs with the scalability and robustness of Transformers, offering a scalable solution for netlist representation learning.

\begin{itemize}
    \item \textbf{Architecture}: DeepGate3 utilizes the pre-trained DeepGate2 as the backbone, extracting initial gate-level embeddings as tokens that capture both the relative position and global functionality of logic gates. To further enhance these embeddings, DeepGate3 employs a component known as the \emph{Refine Transformer} (RT), which refines the initial embeddings and captures the intricate long-term correlations between tokens. This process ensures a deeper and more nuanced understanding of circuit dynamics compared to DeepGate2.
    \item \textbf{Pooling Mechanism}: In contrast to other circuit learning models like Gamora \cite{wu2023gamora} and HOGA \cite{deng2024less}, which use average pooling functions to obtain circuit-level embeddings, DeepGate3 introduces a so-called \emph{Pooling Transformer} (PT) block. Specifically, we use a special \texttt{[CLS]} token that aggregates information between the inputs and outputs of a circuit, serving as the graph-level embedding of the circuit and allowing DeepGate3 to detect and emphasize minor yet significant differences across the circuit more effectively than its predecessors.
    \item \textbf{Pre-training Strategy}: DeepGate3 is pre-trained using a comprehensive set of tasks that operate at multiple levels of circuit analysis. At the gate level, it predicts the logic-1 probability under random simulation and assesses pair-wise truth table similarity, consistent with DeepGate2 \cite{shi2023deepgate2}. At the circuit level, the model selects gates randomly and extracts fan-in cones as separate circuit graphs. It then predicts inherent features of individual cones and quantifies similarities between pairs of cones based on their graph-level embeddings. Importantly, all labels used in these pre-training tasks are generated through logic simulation and circuit analysis, without relying on artificial annotations, ensuring the authenticity and applicability of the training process.
\end{itemize}

This structured approach not only enhances the generalization capabilities of DeepGate3 for practical EDA tasks but also boosts its scalability when managing larger datasets. Such advancements enable DeepGate3 to set a new benchmark in circuit representation learning, significantly outperforming existing GNN-based models.

We rigorously evaluate the scalability and generalization abilities of DeepGate3 by systematically expanding the training dataset. Our experimental results demonstrate that increasing the volume of training data consistently enhances the performance of DeepGate3, showcasing its superior ability to scale effectively. Moreover, the model's proficiency is further evidenced by its superior performance on large circuit configurations compared to previous GNN-based models. To comprehensively assess its generalization capabilities, DeepGate3 is also deployed in solving Boolean Satisfiability (SAT) problem, where it continues to demonstrate its robustness and adaptability in real-world EDA scenarios.

\section{Related Work}
\label{Sec:Related}
\subsection{Circuit Representation Learning}

\begin{figure}
    \centering
    \includegraphics[width=0.85\linewidth]{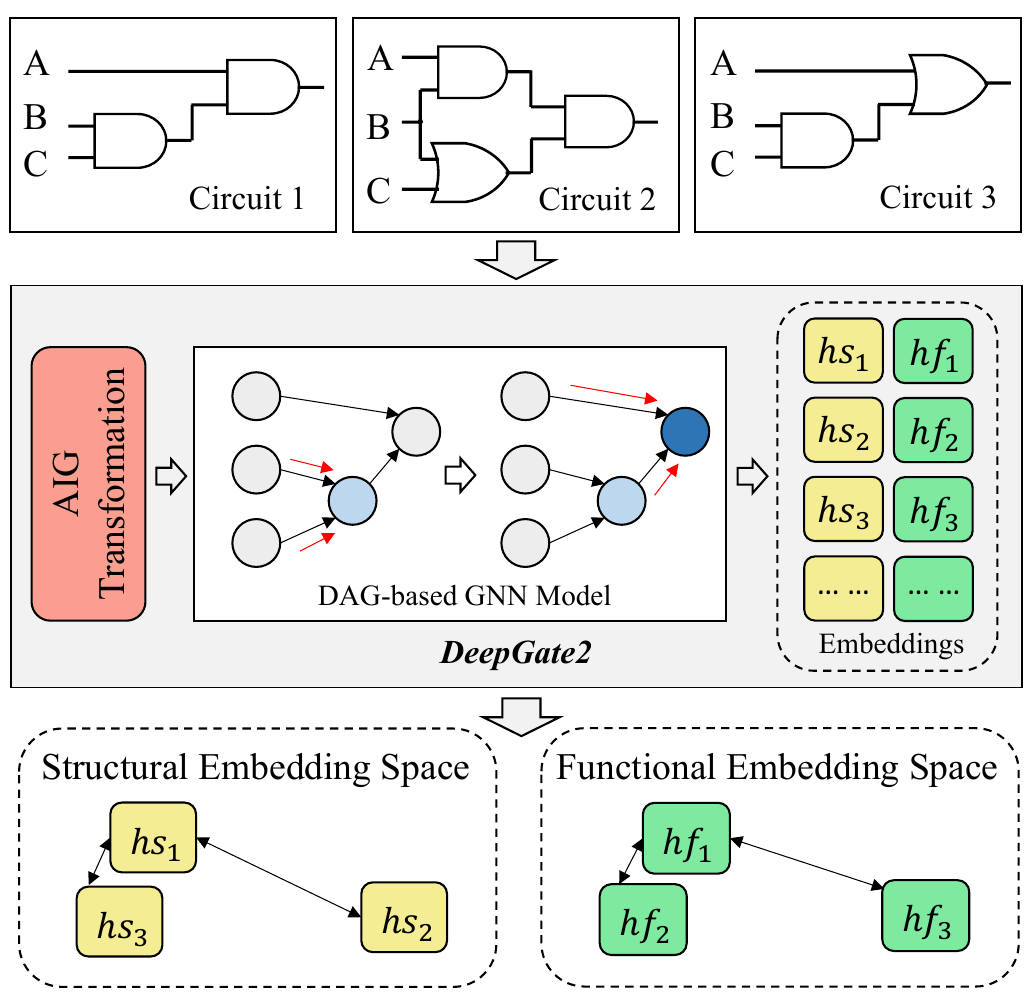}
    \vspace{-10pt}
    \caption{The overview of previous DeepGate2~\cite{shi2023deepgate2}}
    \vspace{-15pt}
    \label{fig:dg2}
\end{figure}

Circuit representation learning is increasingly recognized as a crucial area in the EDA field, reflecting broader trends in AI that emphasize learning general representations for a variety of downstream tasks. Within this context, the DeepGate family~\cite{li2022deepgate, shi2023deepgate2} emerges as a pioneering approach, employing GNNs to map circuit netlists into graph forms and learn gate-level embeddings. The initial version, DeepGate~\cite{li2022deepgate}, focuses on converting arbitrary circuit netlists into And-Inverter Graphs (AIGs), using the logic-1 probability derived from random simulation for model supervision. An evolved iteration, DeepGate2~\cite{shi2023deepgate2}, enhances this model by learning disentangled structural and functional embeddings, enabling it to refine gate-level representation by distinguishing between structural and functional similarities (see Figure~\ref{fig:dg2}), thereby supporting diverse EDA tasks including testability analysis~\cite{shi2022deeptpi}, power estimation~\cite{khan2023deepseq}, and SAT solving~\cite{li2023eda, shi2024eda}.
Recently, the Gamora model~\cite{wu2023gamora} has further advanced the field by showcasing an enhanced reasoning capability through the representation of both logic gates and cones.



Despite these advancements, the application of GNNs in circuit representation poses notable challenges, primarily concerning scalability and generalizability~\cite{liu2024neural}. The HOGA model \cite{deng2024less} attempts to address these issues by introducing a novel message-passing mechanism and an efficient training strategy. However, the inherent limitations of the GNN framework, such as difficulty in managing long-range dependencies and susceptibility to \textit{over-smoothing}~\cite{li2018deeper} and \textit{over-squashing}~\cite{alon2020bottleneck}, continue to restrict the scalability and adaptability of these models~\cite{xu2018powerful}. 

\subsection{Transformer-Based Graph Learning}

Addressing the limitations of GNNs~\cite{kipf2016semi, hamilton2017inductive, velivckovic2017graph}, recent research has pivoted towards incorporating Transformer architectures into graph learning, capitalizing on their renowned ability for handling long-range dependencies through the self-attention mechanism. Graph-BERT~\cite{zhang2020graph} replaces traditional aggregation operators with self-attention to enhance information fidelity across nodes. Similarly, Graphormer~\cite{ying2021transformers} extends the standard Transformer architecture to graph data, demonstrating superior performance across a variety of graph-level prediction tasks. Moreover, GROVER \cite{rong2020self} illustrates the potential of Transformers to capture deep structural and semantic details in molecular graphs. Recent initiatives like OpenGraph~\cite{xia2024opengraph} aim to develop scalable graph foundation models that excel in zero-shot learning tasks across diverse datasets, showcasing the broad applicability of this approach.

Inspired by these advancements, our work explores the integration of Transformer technology into circuit representation learning. DeepGate3 harnesses the strengths of both GNNs and Transformers, aiming to create a robust framework capable of overcoming the traditional limitations of graph learning in the context of circuit netlists. This approach promises not only enhanced scalability and generalization but also superior performance in capturing complex interdependencies within circuit data.



\section{Methodology}
\begin{figure*}[h]
    \centering
    \includegraphics[width=0.8\linewidth]{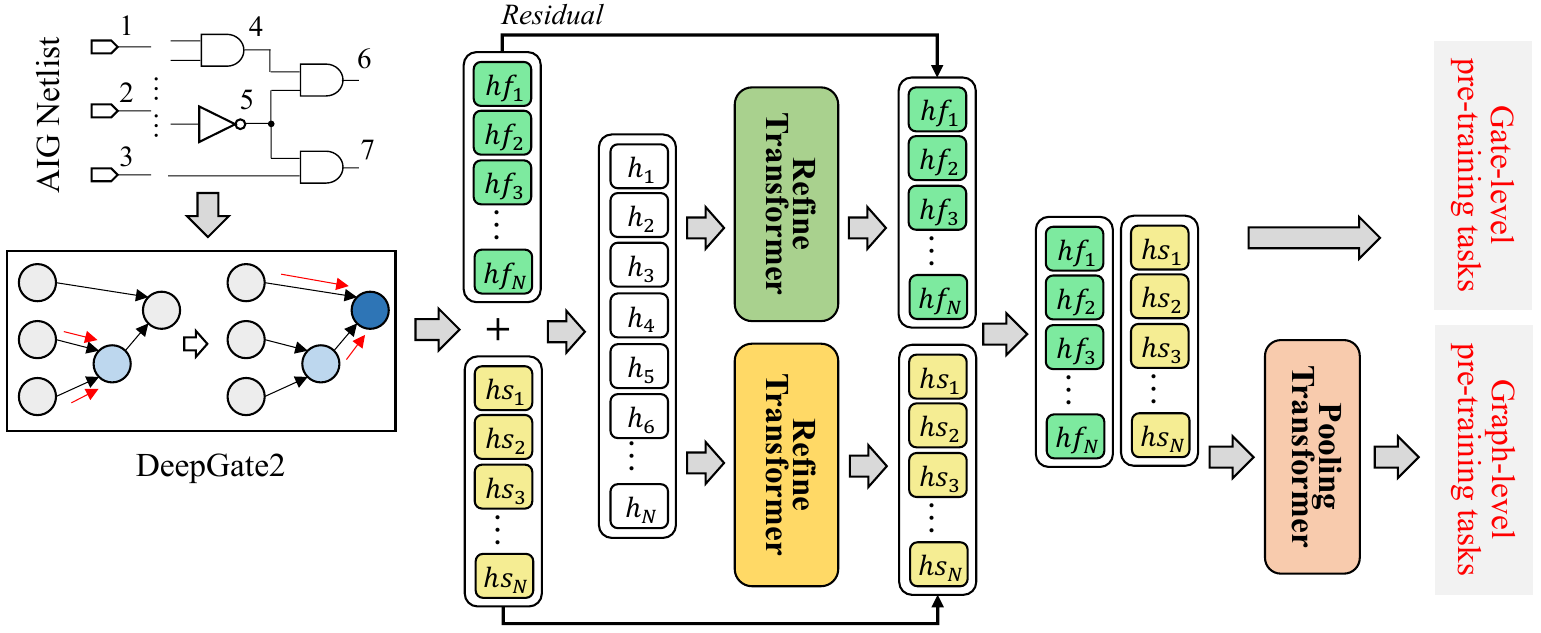}
    \vspace{-5pt}
    \caption{The overview of DeepGate3}
    \vspace{-8pt}
    \label{fig:overview_deepgate3}
\end{figure*}
\label{Sec:Method}
\subsection{Overview of DeepGate3}


The framework of DeepGate3 is illustrated in Figure~\ref{fig:overview_deepgate3}. Initially, the input circuit in And-Inverter Graph (AIG) format is processed using the pre-trained DeepGate2 model, capturing context and structural information for each logic gate (see Section~\ref{Sec:Method:Token}). Next, the gate-level embeddings serving as tokens are processed through the Refine Transformers (RTs). RT improves the embedding sequences by capturing the pair-wise interactions, which is elaborated in Section~\ref{Sec:Method:RT}. Subsequently, we introduce a Transformer-based pooling function in Section~\ref{Sec:Method:Pool}, named Pooling Transformer (PT). PT aggregates information from the refined gate-level embeddings to learn graph-level embeddings. Section~\ref{Sec:Method:Train} provides details on the pre-training tasks. Moreover, to handle large practical circuits, we provide an efficient fine-tuning strategy for DeepGate3 in Section~\ref{Sec:Method:Scale}. 

\subsection{DeepGate2 as Tokenzier} \label{Sec:Method:Token}
The Transformer-based circuit learning model requires the tokenization of the circuit graph into a sequence. However, directly converting the circuit into a sequence of gates solely based on their gate type and topological order leads to a loss of valuable information. This limitation is particularly evident in AIG, where there are only two gate types. Despite the limited diversity of gate types, each gate holds valuable contextual information, highlighting the need for a domain-specific tokenization approach. 

We propose to utilize DeepGate2 as a tokenizer, which embeds distinct spatial and functional features of logic gates. However, the original DeepGate2 assumes that each PI has a $0.5$ probability of being logic-1, which is inadequate for handling circuits under arbitrary workloads. We re-train the DeepGate2 model with specific workloads, i.e., we randomly initialize the probability $p_i$ of each PI $i$. Therefore, such mechanism allows model to learn workload-aware embeddings, enhancing its ability to provide more informative and adaptable representations.
Aside from this modification, DeepGate2 retains its original setting.

The input circuit graph is denoted as $\mathcal{G} = \langle \mathcal{V}, \mathcal{E} \rangle$, where $\mathcal{V} = [v_1, v_2, \ldots, v_n]$ represents the set of vertices, and $\mathcal{E}$ represents the set of edges. To initialize the functional embeddings of gate $i$, where $i$ belongs to PI $\mathcal{I}$, we assign a repeated value of $p_i$ to each dimension for a total of $d$ times, i.e., $hf_i = Repeat(p_i, d), \ i \in \mathcal{I}$

Next, the structural embeddings of PIs are assigned a set of randomly generated but orthogonal vectors. The remaining gate embeddings are all initialized to zero. Formally, DeepGate2 tokenizes the circuit graph $\mathcal{G}$ into two sequences of gate-level token embeddings: structural embeddings $HS$ and functional embeddings $HF$ as follows:
\begin{equation}
    \begin{split}
        HS, HF & = DeepGate2(\mathcal{G}, \mathbf{p}), \\
        HS &\Leftrightarrow [hs_1, hs_2, ..., hs_n] \\ 
        HF &\Leftrightarrow [hf_1, hf_2, ..., hf_n]
        \vspace{-10pt}
    \end{split}
\end{equation}



\begin{figure}[!t]
    \centering
    \includegraphics[width=0.9\linewidth]{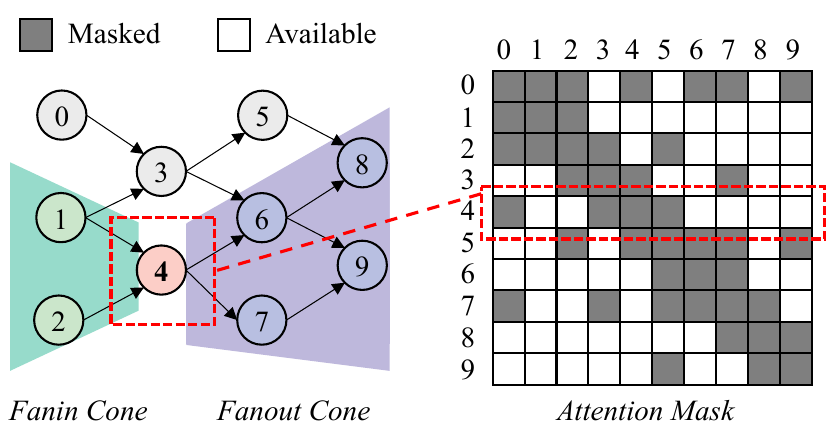}
    \vspace{-5pt}
    \caption{The example of fanin-fanout cones and the corresponding mask}
    \vspace{-10pt}
    \label{fig:cone}
\end{figure}

\subsection{Refine Transformer} \label{Sec:Method:RT}
\textbf{Positional embedding. } Given initialized functional embedding sequence $HF=[hf_1,hf_2,...,hf_n]$ and structural embedding $HS=[hs_1,hs_2,...,hs_n]$ sequence acquired by DeepGate2, we propose to use two independent Transformers to get refined functional embedding and structural embedding, respectively. However, a problem lies: \textit{How to design a positional embedding to represent the order information of the gate embedding sequence?} 

Different from data like natural language, which are inherently sequential, the gates in AIGs does not have a clear order. Existing graph sequentialization methods like topological order and canonical order cannot differentiate the isomorphic graphs~\cite{dong2022pace}, making it impractical to rely on explicit order as positional embeddings. To tackle this problem, we propose to use implicit embedding as positional embedding. 
\begin{itemize}
    \item For functional embedding, we use corresponding structural embedding as positional embedding to identify AIGs with the same function but different structure, e.g. Circuit 1 and Circuit 2 in Figure~\ref{fig:dg2}.
    \item  Similarly, for structural embedding, we use corresponding functional embedding as positional embedding. This can also help identify AIGs with similar structure but greatly different function, as illustrated by the example of Circuit 1 and Circuit 3 in Figure~\ref{fig:dg2}. 
\end{itemize}

\noindent\textbf{Fanin-fanout Cone.} While the Transformer model has the capability to aggregate global information from an AIG, it should be denoted that not all the information is equally useful for the representation learning. As shown in Figure~\ref{fig:cone}, the functionality of Gate 4 is directly impacted by its predecessors, namely Gate 1 and Gate 3, while Gates 6 to 9 are influenced by Gate 4. Conversely, the remaining Gate 0, 3, 5 neither affect nor are affected by the gate under consideration. Subsequently, we identify the fanin and fanout cones for each gate. For Gate 4, we create the Attention Mask matrix in Figure~\ref{fig:cone}, following the masking approach in Transformer~\cite{vaswani2017attention}. Within this matrix $MASK$, "Available" denotes the corresponding tokens will participate in attention computation, whereas "Masked" means the opposite. In other words, only the gates within the fanin or fanout cones are involved in the attention computation process. 

\noindent\textbf{Transformer Architecture.} As shown in Figure~\ref{fig:overview_deepgate3}, DeepGate3 includes two independent Refine Transformers (denoted as $RT_{s}$ and $RT_{f}$) for learning functional and structural information, respectively. As described before, we use $HS$ and $HF$ as the positional embedding. Therefore, we get the refined embedding as follows:
\begin{equation}
    \begin{split}
        \widetilde{HF} &= RT_{f}(HF+HS, MASK)\\
        \widetilde{HS} &= RT_{s}(HS+HF, MASK) 
    \end{split}
\end{equation}

\begin{figure}[!t]
    \centering
    \includegraphics[width=1.\linewidth]{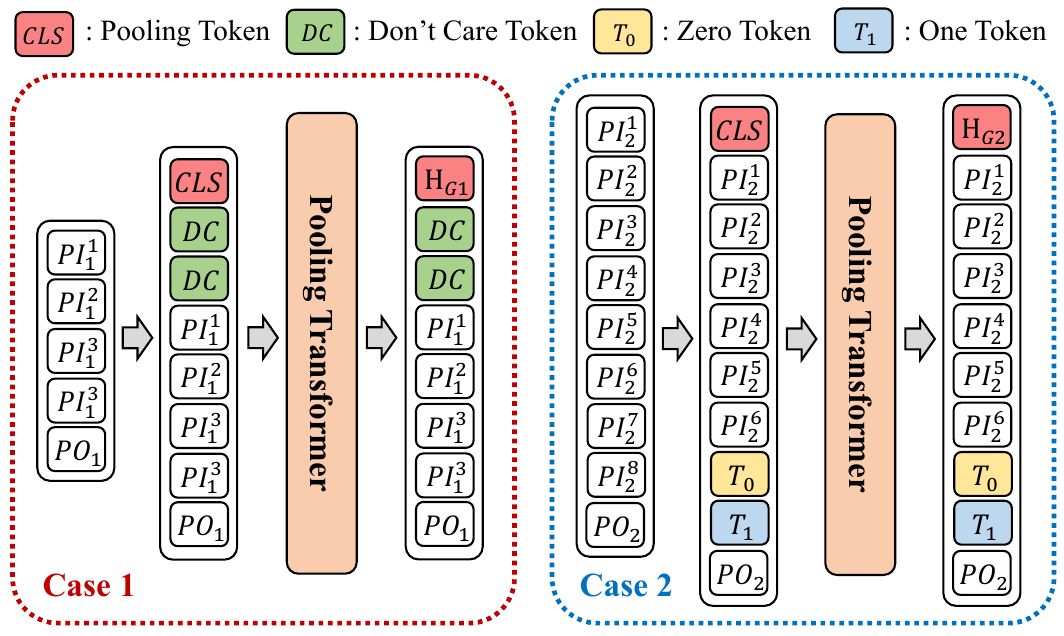}
    \vspace{-5pt}
    \caption{The overview of functional pooling Transformer}
    \vspace{-10pt}
    \label{fig:pooling_transformer}
\end{figure}

\subsection{Pooling Transformer} \label{Sec:Method:Pool}
DeepGate3 aims to learn not only gate level embeddings but also sub-graph level embeddings. To obtain sub-graph level embeddings, we utilize a lightweight Transformer, referred to as the Pooling Transformer (PT). 

We first extract a set of sub-graphs $\mathcal{S}$ from the circuits based on $l$-hop cones. The construction of a sub-graph involves randomly selecting a logic gate within the circuit and extracting its $l$-order predecessors to form the sub-graph. The resulting $l$-hop sub-graph encompasses a maximum of $2^{l-1}$ PIs and $2^{l}-1$ gates. In our default settings, we assign number of hops as $l=4$. It is worth noting that the order of the PIs can influence the truth table of a sub-graph. To ensure uniformity, we reorder the indexes of nodes based on the canonical labels generated by Nauty tool~\cite{mckay1981practical}. 

\noindent \textbf{Functional Pooling Transformer. } For AIGs, the function of a $l$-hop cone is directly linked to its truth table. Therefore, we design a pre-training task that predicts the truth table of a sub-graph using only its PIs and PO. However, the variable number of PIs in a sub-graph and the fixed length of the truth table remain a challenge. During the model pre-training, we fix the truth table length to $64$, which corresponds to a sub-circuit with $6$ inputs and $1$ output. During the model inference, PT can process circuits with any number of PIs and POs.

To handle sub-graphs with different numbers of inputs, we introduce three special learnable tokens: \texttt{[Don't care]}, \texttt{[Zero]}, and \texttt{[One]}. These tokens help adjust the length of the truth table as needed. As illustrated in Figure~\ref{fig:pooling_transformer}, adjustments are as follows:
\begin{itemize}
    \item Case 1. When there are fewer than 6 PIs, \texttt{[Don't care]} tokens are added to increase the count to 6 and expand the truth table accordingly.
    \item Case 2. When there are more than 6 PIs, some PIs are replaced randomly with \texttt{[Zero]} or \texttt{[One]} to adjust the truth table.
\end{itemize}
Similar to NLP techniques, we introduce a special learnable token \texttt{[CLS]} at the beginning of the embedding sequence to aggregate information from a sub-graph. The order of PIs is crucial for predicting the truth table, so we use learnable positional embeddings, following the method used by BERT~\cite{devlin2018bert}.

\noindent \textbf{Structural Pooling Transformer. } Unlike the functional counterpart, the Structural Pooling Transformer focuses on the structural information which is related to all gates within a sub-graph. Therefore, structural Pooling Transformer leverages all gates in a sub-graph, along with the \texttt{[CLS]} token, to obtain a structural pooling embedding. Similar to functional pooling Transformer, the structural version also uses learnable positional embeddings.

\subsection{Model Pre-training} \label{Sec:Method:Train}
\noindent\textbf{Gate-level Pre-training Tasks.}
DeepGate3 model is pre-trained with a series of tasks in both gate-level and graph-level. To disentangle the functional and structural embeddings, we employ pre-training tasks with different labels to supervise the corresponding components. 

Regarding function-related tasks at the gate-level, we incorporate the pre-training tasks from DeepGate2, which involve predicting the logic-1 probability of gates and predicting the pair-wise truth table distance. We sample the gate pairs $\mathcal{N}_{gate\_tt}$ and record the corresponding simulation response as incomplete truth table $T_{i}$. The pair-wise truth table distance $D^{gate\_tt}$ is calculated using the following formula:
\begin{equation}
    D^{gate\_tt}_{(i,j)} = \frac{HammingDistance(T_i, T_j)}{length(T_i)}, (i, j)\in \mathcal{N}_{gate\_tt}
\label{eq:tt_distance}
\end{equation}

The loss functions for gate-level functional pre-training tasks are depicted as below. In DeepGate3, multiple Multi-Layer Perceptron (MLP) heads are utilized to readout embeddings.
\begin{equation} \label{Eq:loss:gatefunc}
    \begin{split}
        L_{gate}^{prob} & = L1Loss(p_k, MLP_{prob}(hf_k)), k \in \mathcal{V} \\
        L_{gate}^{tt\_pair} & = L1Loss(D^{gate\_tt}_{(i, j)}, MLP_{gate\_tt}(hf_i, hf_j)), (i, j) \in \mathcal{N}_{gate\_tt}
    \end{split}
\end{equation}

In addition, we incorporate supervisions for structural learning by predicting logic levels and the pair-wise connections. The prediction of pairwise connections is treated as a classification task, where a sampled gate pair $(i, j) \in \mathcal{N}_{gate\_con}$ can be classified into three categories: gate $i$ can propagate to gate $j$ across edges, gate $j$ can propagate to gate $i$ across edges, or gate $i$ cannot reach gate $j$ at any time. We list the loss functions as below, where $Lev_k$ is the logic level of gate $k$. 
\begin{equation} \label{Eq:loss:gatestru}
    \begin{split}
        L_{gate}^{lev} & = L1Loss(Lev_k, MLP_{lev}(hs_k)), k \in \mathcal{V} \\
        L_{gate}^{con} & = BCELoss(MLP_{con}(hs_i, hs_j)), (i, j) \in \mathcal{N}_{gate\_con}
    \end{split}
\end{equation}

\noindent\textbf{Graph-level Pre-training Tasks.}
To ensure an adequate quantity of high-quality training samples, we construct graph-level supervisions using the sub-graphs extracted from the original netlists, i.e., the extracted $l$-hop sub-graphs elaborated in Section~\ref{Sec:Method:Pool}. 

For the individual sub-graph, we prepare the truth table by complete simulation, denoted as $T_s$. Besides, we collect two structural-related characteristics of each sub-graph, namely the number of nodes $Size(s)$ and the depth $Depth(s)$. After obtaining the functional embedding $hf^s$ and structural embedding $hs^s$ by pooling Transformer, we can supervise the model training using the following loss functions, where $s \in \mathcal{S}$. 
\begin{equation} \label{Eq:loss:graph}
    \begin{split}
        L_{graph}^{size} & = L1Loss(Size(s), MLP_{size}(hs^s)) \\ 
        L_{graph}^{depth} & = L1Loss(Depth(s), MLP_{depth}(hs^s)) \\ 
        L_{graph}^{tt} & = BCELoss(T_s, MLP_{tt}(hf^s)) 
    \end{split}
\end{equation}

Moreover, we introduce loss functions to effectively capture the pair-wise correlations between sub-graphs. We leverage Eq.~\eqref{Eq:loss:graphpair} to predict both the truth table distance $D_{(s_1, s_2)}^{graph\_tt}$ and the graph edit distance~\cite{bunke1997relation} $D_{(s_1, s_2)}^{graph\_ged}$ between two sub-graphs ($s_1, s_2$), which indicate the functional and structural relationships, respectively. 
\begin{equation} \label{Eq:loss:graphpair}
    \begin{split}
        D_{(s_1, s_2)}^{graph\_tt} & = \frac{HammingDistance(T_{s_1}, T_{s_2})}{length(T_{s_1})} \\
        L_{graph}^{tt\_pair} & = L1Loss(D_{(s_1, s_2)}^{graph\_tt}, MLP_{graph\_tt}(hf^{s_1}, hf^{s_2})) \\ 
        D_{(s_1, s_2)}^{graph\_ged} & = GraphEditDistance(s_1, s_2) \\ 
        L_{graph}^{ged\_pair} & = L1Loss(D_{(s_1, s_2)}^{graph\_ged}, MLP_{graph\_ged}(hs^{s_1}, hs^{s_2})) \\
    \end{split}
\end{equation}

To establish a connection between the embedding spaces of gate-level and graph-level representations, we enable the model to determine whether gate $k$ belongs to sub-graph $s$ according to the structural embeddings. 
\begin{equation} \label{Eq:loss:in}
    L_{in} = BCELoss(\{0, 1\}, MLP_{in}(hs_k, hs^{s}))
\end{equation}

The overall loss $Loss_{all}$ in pre-training sums up all $10$ loss function values in Eq.~\eqref{Eq:loss:gatefunc}~\eqref{Eq:loss:gatestru}~\eqref{Eq:loss:graph}~\eqref{Eq:loss:graphpair} and~\eqref{Eq:loss:in}.
\begin{equation} \label{Eq:loss:all}
    \begin{split}
        Loss_{all} & = L_{gate}^{prob} + L_{gate}^{tt\_pair} + L_{gate}^{lev} + L_{gate}^{con} + L_{graph}^{size} + L_{graph}^{depth} \\ 
        & + L_{graph}^{tt} + L_{graph}^{tt\_pair} + L_{graph}^{ged\_pair} + L_{in}
    \end{split}
\end{equation}

\begin{figure}[!t]
    \centering
    \includegraphics[width=1.0\linewidth]{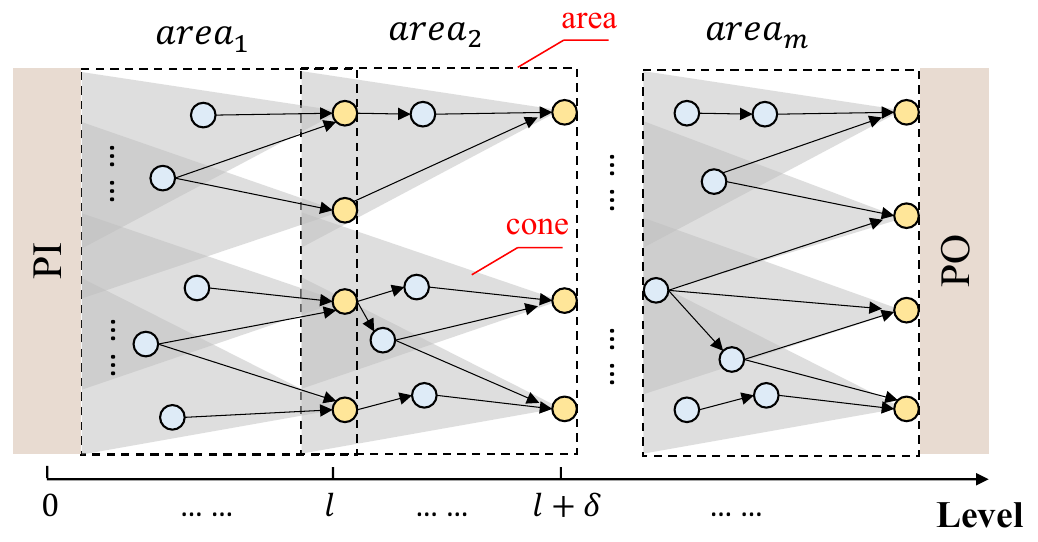}
    \vspace{-15pt}
    \caption{The partitioned area of large AIG}
    \vspace{-15pt}
    \label{fig:area}
\end{figure}

\subsection{Fine-tuning on Large AIGs} \label{Sec:Method:Scale}
Since the computational complexity of self-attention mechanism grows quadratically with the sequence length, there is a practical limitation on the maximum sequence length in Transformer block. However, real circuits typically contain a large number of gates, far exceeding the maximum sequence length of $512$ in our Refine Transformer. One approach to address this limitation is to sparsify the attention matrix by restricting the field of view to fixed, predefined patterns such as local windows and block patterns of fixed strides~\cite{efficient_transformer_survey,efficient_transformer1,efficient_transformer2,efficient_transformer3,efficient_transformer4,efficient_transformer5}. 
Similarly, we introduce a window-shifting method to facilitate the fine-tuning of the model on the large AIG. Specifically, RT calculates the self-attention of the nodes within a shifting window. This shifting window slides by one step at each iteration, allowing the RT to capture the entire large AIG. 

\noindent\textbf{Large Circuit Partition. } The partitioned circuits covered by the shifting window are shown in Figure~\ref{fig:area}. Initially, we focus on gathering all the $l$-hop \textit{cones} that terminate in logic level $l$. These cones are subsequently merged, forming an \textit{area} denoted as $area_1$. Moving forward, we continue collecting and merging cones with output gates situated in level $l+\delta$. Here, $\delta$ represents the level gap between two distinct areas. It is important to note that the chosen value of $\delta$ must be smaller than $l$ in order to guarantee an overlap between the two areas. The aforementioned process is repeated iteratively until the partitioned areas cover the entire circuit. Furthermore, we impose a constraint wherein the maximum number of gates within an area is limited to $512$. If the number of gates in any given area exceeds the limitation, it will be further divided into small areas. 

\noindent\textbf{Window-Shifting Pipeline. } As depicted in Algorithm~\ref{Alg:hopformer}, given an AIG with the partitioned area $\{area_1, area_2, \ldots, area_m\}$, we first tokenize the entire AIG into embedding sequences $HF$ and $HS$. 
Then, we acquire the embedding sequences $\{Hf_{1}, Hf_{2}, $ $\ldots, Hf_{m}\}$ and $\{Hs_{1}, Hs_{2}, \ldots, Hs_{m}\}$ for each area. The embeddings in area $area_t$ covering by the shifting window are then fed into the pre-trained Refine Transformer to obtain the refined embeddings $\widetilde{H}f_t$ and $\widetilde{H}s_t$. Subsequently, the window move forward to the next area $area_{t+1}$. As there is overlap among the areas, we adopt an average updating strategy, computing the embedding of an overlapping gate as the average of the corresponding gate embeddings. 
Finally, these embeddings are passed to the fine-tuning task-specific heads for various downstream tasks.

\begin{algorithm}
	\renewcommand{\algorithmicrequire}{\textbf{Input:}}
	\renewcommand{\algorithmicensure}{\textbf{Output:}}
	\caption{Large circuit encoding for fine-tuning}
	\label{Alg:hopformer}
	\begin{algorithmic}[1]
		\REQUIRE \  \\Pre-trained Refine Transformer $RT_{f}$,$RT_{s}$, 
        \\ Pre-trained DeepGate2 $DeepGate2$,
        \\ Large AIG $\mathcal{G}$ with PI workload $\mathbf{p}$,
        \\  Partition of Large AIG $\{area_1, area_2,..., area_m\}$,
        
        \STATE $HF,HS \gets DeepGate2(\mathcal{G}, \mathbf{p})$
        
		\FOR{$t$ in $[1, 2, ..., m]$}
        \STATE Acquire $Hf_{t}$, $Hs_{t}$ in $HF, HS$ according to the index of $area_t$ in large AIG $\mathcal{G}$
        \STATE Compute $MASK$ of $area_t$
        \STATE $\widetilde{H}f_t \gets RT_{f}(Hf_t+Hs_t, MASK)$ 
        \STATE $\widetilde{H}s_t \gets RT_{s}(Hs_t+Hf_t, MASK)$
        \STATE $HF \gets average\_update(\widetilde{H}f_t)$
        \STATE $HS \gets average\_update(\widetilde{H}s_t)$
		\ENDFOR
		\STATE \textbf{return} Refined embeddings $HF$, $HS$
	\end{algorithmic}  
\end{algorithm}
\vspace{-10pt}

\section{Experiments}
\label{Sec:Experiment}

\begin{table*}[]
\caption{The comparison of scalability between DeepGate2 and DeepGate3}
\label{tab:scaling}
\vspace{-5pt}
\centering
\setlength\tabcolsep{3pt}
\begin{tabular}{@{}c|c|ccccc|cccccccc|c@{}}
\toprule
\multirow{2}{*}{Model}     & \multirow{2}{*}{Data} & \multicolumn{5}{c|}{Gate-level Tasks}                                                                    & \multicolumn{8}{c|}{Graph-level Tasks}                                                                                                                                      & \multirow{2}{*}{$L_{all}$} \\ \cmidrule(lr){3-15}
                           &                       & $L_{gate}^{prob} $ & $L_{gate}^{tt\_pair}$ & $L_{gate}^{lev} $ & $L_{gate}^{con} $ & $P^{con} $        & $L_{graph}^{tt} $ & $P^{tt} $        & $L_{graph}^{tt\_pair} $ & $L_{graph}^{ged\_pair} $ & $L_{graph}^{size} $ & $L_{graph}^{depth} $ & $L_{in} $        & $P^{in} $         &                               \\ \midrule
\multirow{6}{*}{\rotatebox{90}{DeepGate2}} & 1\%                   & 0.0731            & 0.0294                & 0.3780           & 0.3635           & 84.97\%          & 0.2904           & 0.1423          & 0.1312                 & 0.0841                  & 0.9436             & 0.3132              & 0.5498          & 69.98\%          & 3.1563                        \\
                           & 5\%                   & 0.0193            & 0.0172                & 0.7454           & 0.3482           & \textbf{85.62\%} & 0.2094           & 0.1057          & 0.1051                 & 0.0768                  & 0.8570             & 0.2109              & 0.5013          & 74.13\%          & 3.0906                        \\
                           & 10\%                  & 0.0195            & 0.0217                & 0.3252           & 0.3559           & 84.91\%          & 0.2121           & 0.1072          & 0.1040                 & 0.0773                  & 0.7877             & 0.2264              & 0.4952          & 74.47\%          & 2.6250                        \\
                           & 30\%                  & \textbf{0.0083}   & 0.0142                & \textbf{0.1729}  & 0.3479           & 85.42\%          & 0.1913           & 0.0970          & 0.0924                 & 0.0784                  & 0.7695             & 0.2529              & 0.4753          & 75.81\%          & 2.4031                        \\
                           & 50\%                  & 0.0159            & 0.0136                & 0.3178           & \textbf{0.3456}  & 84.95\%          & 0.1737           & 0.0914          & \textbf{0.0756}        & 0.0710                  & \textbf{0.5901}    & 0.1148              & 0.4310          & 78.52\%          & \textbf{2.1491}               \\
                           & 100\%                 & 0.0131            & \textbf{0.0107}       & 0.4041           & 0.3384           & 85.58\%          & \textbf{0.1703}  & \textbf{0.0898} & \textbf{0.0756}        & \textbf{0.0708}         & 0.5965             & \textbf{0.1122}     & \textbf{0.4298} & \textbf{78.57\%} & 2.2215                        \\ \midrule
\multirow{6}{*}{\rotatebox{90}{DeepGate3}} & 1\%                   & 0.0847            & 0.0573                & 3.3150           & 0.3919           & 84.82\%          & 0.3160           & 0.1387          & 0.0913                 & 0.0985                  & 5.0383             & 0.7082              & 0.5802          & 66.71\%          & 10.6814                       \\
                           & 5\%                   & 0.0388            & 0.0269                & 0.3526           & 0.2668           & 89.31\%          & 0.2565           & 0.0931          & 0.0701                 & 0.0665                  & 0.3401             & 0.1314              & 0.3560          & 83.79\%          & 1.9057                        \\
                           & 10\%                  & 0.0310            & 0.0220                & 0.3373           & 0.2367           & 89.99\%          & 0.1967           & 0.0710          & 0.0652                 & 0.0616                  & 0.2518             & 0.1138              & 0.3291          & 85.04\%          & 1.6452                        \\
                           & 30\%                  & 0.0175            & 0.0143                & \textbf{0.2526}  & 0.1856           & 91.94\%          & 0.1826           & 0.0459          & 0.0485                 & 0.0549                  & 0.1270             & 0.0562              & 0.2580          & 88.45\%          & 1.1972                        \\
                           & 50\%                  & 0.0156            & 0.0109                & 0.3335           & 0.1552           & 92.99\%          & 0.1787           & 0.0348          & 0.0405                 & 0.0470                  & 0.1120             & \textbf{0.0236}     & 0.1524          & 94.21\%          & 1.0694                        \\
                           & 100\%                 & \textbf{0.0136}   & \textbf{0.0082}       & 0.2672           & \textbf{0.1480}  & \textbf{93.32\%} & \textbf{0.0862}  & \textbf{0.0237} & \textbf{0.0387}        & \textbf{0.0457}         & \textbf{0.0888}    & 0.0245              & \textbf{0.1344} & \textbf{95.18\%} & \textbf{0.8553}               \\ \bottomrule
\end{tabular}
\end{table*}

\subsection{Data Preparation}
We collect the circuits for pre-training from various sources, including benchmark netlists in ITC99~\cite{ITC99}, IWLS05~\cite{albrecht2005iwls}, EPFL~\cite{EPFLBenchmarks}, and synthesizable register-transfer level (RTL) designs from OpenCore~\cite{takeda2008opencore} and Github~\cite{thakur2023benchmarking}. All designs are transformed into AIG netlists by ABC tool~\cite{brayton2010abc}. If the original netlists are too large, we random partition them into small circuits, ensuring that the maximum number of nodes in each circuit remains below $512$. Furthermore, we augment the dataset by synthesizing netlists using various logic synthesis recipes. The statistical details of our pre-training dataset can be found in Table~\ref{TAB:Data}, which comprises a total of $67,905$ circuits. Note that we pre-train DeepGate3 on these small circuits, and for downstream tasks, we transfer DeepGate3 to large circuits.

\begin{table}[!t]
\centering
\caption{The statistics of pre-training dataset} \label{TAB:Data}
    \vspace{-5pt}
\begin{tabular}{@{}ll|ll|ll@{}}
\toprule
\multicolumn{1}{c}{\multirow{2}{*}{Source}} & \multicolumn{1}{c|}{\multirow{2}{*}{\# Circuits}} & \multicolumn{2}{c|}{\# Nodes}     & \multicolumn{2}{c}{\# Levs}     \\
\multicolumn{1}{c}{}                        & \multicolumn{1}{c|}{}                             & Avg.            & Std.            & Avg.           & Std.           \\ \midrule
ITC99                                       & 15,991                                            & 268.83          & 99.06           & 31.97          & 16.89          \\
IWLS05                                      & 4,233                                             & 187.87          & 93.51           & 17.66          & 7.80           \\
EPFL                                        & 10,478                                            & 279.40          & 106.40          & 27.55          & 15.09          \\
OpenCore                                    & 30,582                                            & 184.90          & 77.69           & 17.56          & 10.92          \\
Github                                      & 6,621                                             & 245.03          & 133.72          & 20.41          & 13.17          \\ \midrule
\textbf{Total}                              & \textbf{67,905}                                   & \textbf{225.29} & \textbf{104.03} & \textbf{22.78} & \textbf{14.65} \\ \bottomrule
\end{tabular}
    \vspace{-10pt}
\end{table}

\subsection{Implementation Details}
The dimensions of both the structural embedding $h_s$ and the functional embedding $h_f$ are set to $128$. The depth of Refine Transformer is $12$ and the depth of Pooling Transformer is $3$. All pre-training task heads are 3-layer multilayer perceptrons (MLPs). For pair-wise tasks in Section~\ref{Sec:Method:Train}, we concatenate their embeddings as the input to the pre-training task head.

We pre-train our model with tasks described in Section~\ref{Sec:Method:Train} for $500$ epochs to ensure convergence. This pre-training is performed with a batch size of $128$ on 8 Nvidia A800 GPUs. We utilize the Adam optimizer with a learning rate of $10^{-4}$.

\subsection{Evaluation Metric}
\label{sec:eval_metric}
To better assess the performance of our model, we utilize several metrics as follows.

\noindent\textbf{Overall Loss. } To calculate the overall loss $L_{all}$, we simply sum up all the losses as Eq.~\eqref{Eq:loss:all}.

\noindent\textbf{Error of Truth Table Prediction. } Given the 6-input sub-graph $s$ in test dataset $\mathcal{S'}$ with special token in Section~\ref{Sec:Method:Pool}, we predict the 64-bit truth table based on the graph-level functional embedding $hf^s$. The prediction error is calculated by the Hamming distance between the prediction and ground truth. 
\begin{equation}
    P^{tt} = \frac{1}{len(\mathcal{S'})} \sum_{s}^{\mathcal{S'}} HammingDistance(T_s, MLP_{tt}(hf^s))
\end{equation}

\noindent\textbf{Accuracy of Gate Connection Prediction. } Given the structural embedding of the gate pair $(i, j)$ in test dataset $\mathcal{N}_{con}'$ and binary label $y^{con}_{(i, j)} = \{0, 1, 2\}$, we predict the gate connection and define the accuracy as follows:
\begin{equation}
    P^{con} = \frac{1}{len(\mathcal{N}_{con}')}\sum_{(i, j)}^{\mathcal{N}_{con}'}\mathbbm{1}(y^{con}_{(i, j)}, MLP_{con}(hs_i, hs_j))
\end{equation}

\noindent\textbf{Accuracy of Gate-in-Graph Prediction. } Given the gate-graph pair $(k, s)$ in test dataset $\mathcal{N}_{in}'$, we predict whether the gate is included by the sub-graph with the gate structural embedding $hs_k$ and the sub-graph structural embedding $hs^{s}$. The binary label is noted as $y^{in}_{(k, s)} = \{0 ,1\}$. 
\begin{equation}
    P^{in} = \frac{1}{len(\mathcal{N}_{in}')}\sum_{(k, s)}^{\mathcal{N}_{in}'} \mathbbm{1}(MLP_{in}(hs_k, hs^{s}),y_k^{on})
\end{equation}

\subsection{Data Scalability of DeepGate3}

\begin{figure*}
\centering
\subfloat[Overall Loss]{
\begin{minipage}[t]{0.24\linewidth}
    \centering
    \includegraphics[width=1.0\textwidth]{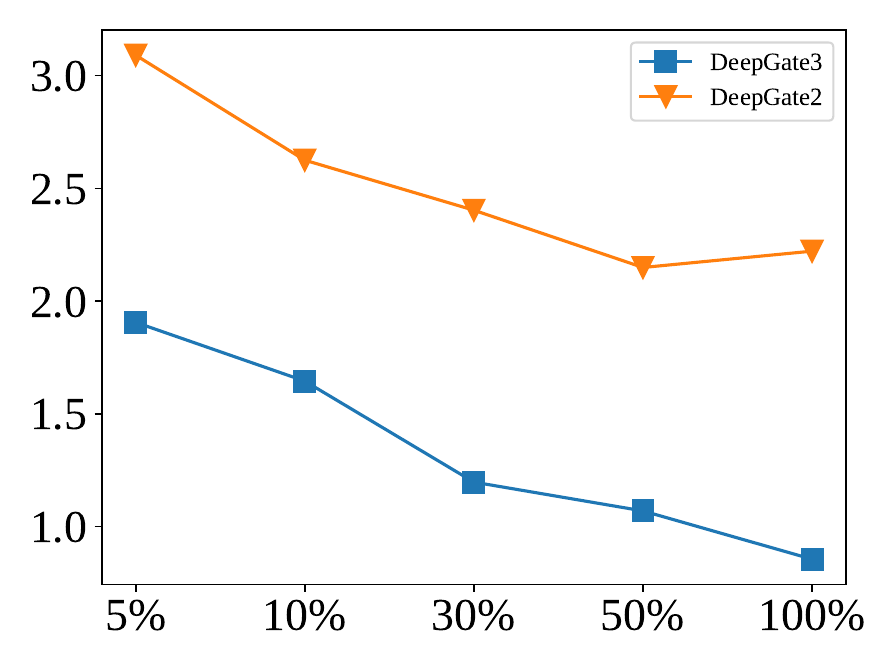}
\end{minipage}
\label{fig:overall_loss}
}
\subfloat[Truth Table Prediction]{
\begin{minipage}[t]{0.24\linewidth}
    \centering
    \includegraphics[width=1.\textwidth]{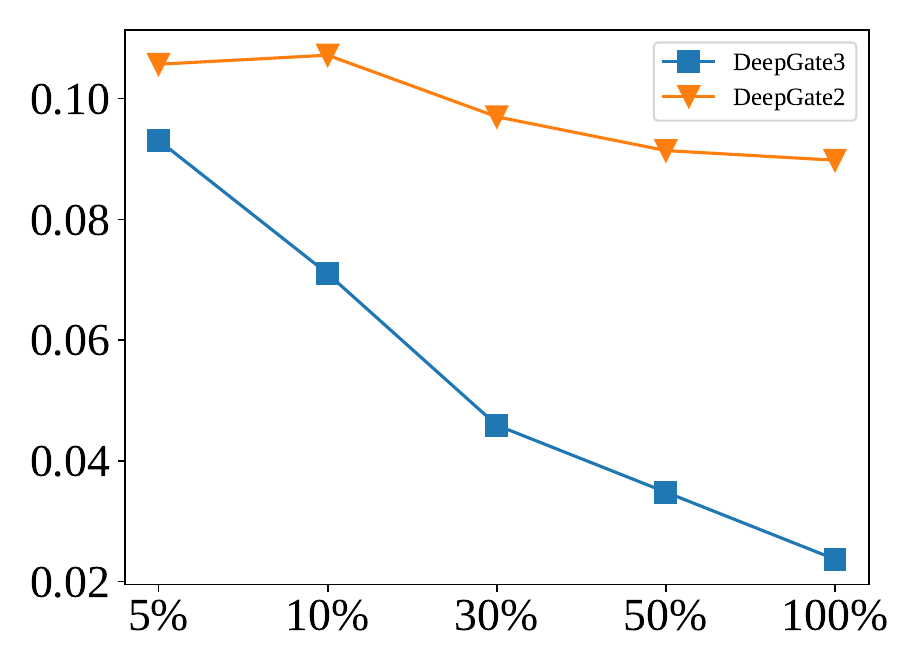}
\end{minipage}
\label{fig:tt_pre}
}
\subfloat[Gate Connection Prediction]{
\begin{minipage}[t]{0.24\linewidth}
    \centering
    \includegraphics[width=1.\textwidth]{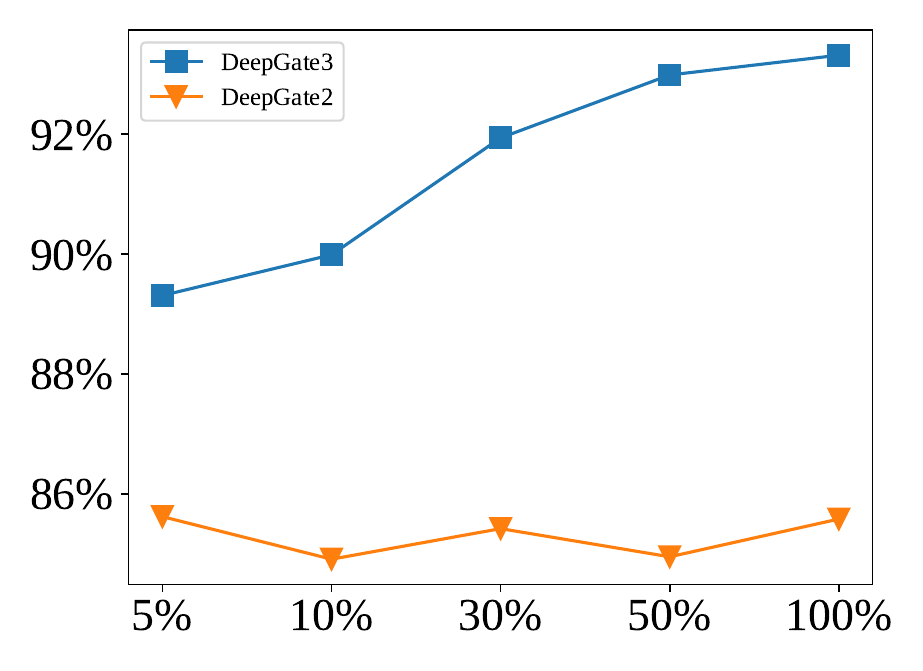}
\end{minipage}
\label{fig:con_pre}
}
\subfloat[Gate-in-Graph Prediction]{
\begin{minipage}[t]{0.24\linewidth}
    \centering
    \includegraphics[width=1.\textwidth]{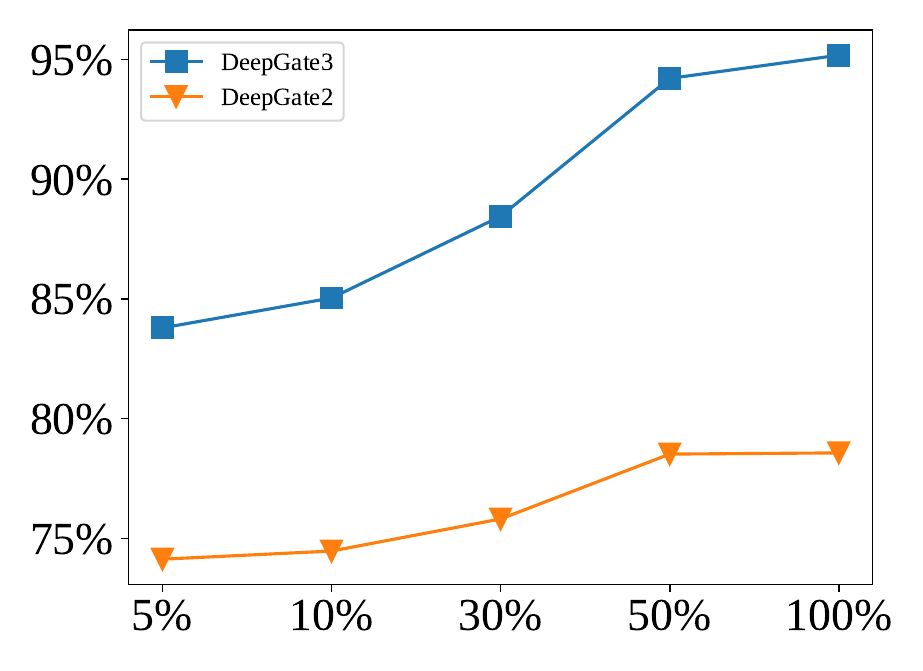}
\end{minipage}
\label{fig:Gate-in-Graph}
}
  \caption{The comparing of Overall Loss, Error of Truth Table Prediction, Accuracy of Gate Connection Prediction and Accuracy of Gate-in-Graph Prediction loss among 5\%, 10\%, 30\%, 50\%, 100\% data.}
  \label{fig:scaling}
\end{figure*}

The data scalability, indicating that the model performance continues to increase as the training data expands, has been demonstrated across various models~\cite{scaling1,scaling2,scaling3,scaling4}. In this section, we conduct experiments to investigate the data scalability of DeepGate3. We partition our original training dataset, consisting of $67,905$ netlists, into subsets of $1\%$, $5\%$, $10\%$, $30\%$, and $50\%$ of the original size ($100\%$). We train multiple models with the same settings on each scaled training dataset and train them for $500$ epochs to ensure the loss converges. All the models are validated with the same testing dataset to ensure the fairness. The experimental results with all loss values and evaluation metrics are shown in Table~\ref{tab:scaling}. 

To further investigate the data scalability of DeepGate3, we train DeepGate2 on these subsets of train datasets with the same hyperparameters. The corresponding results are presented in Table~\ref{tab:scaling}. Furthermore, to provide a clear visualization of the results, we depict the performance metrics in Figure~\ref{fig:scaling}, including the overall loss, the error of truth table prediction, the accuracy of gate connection prediction, and the accuracy of gate-in-graph prediction. It is worth noting that the $1\%$ subset of the training dataset contains only $679$ netlists, resulting in ineffective performance for all models trained with this subset. Therefore, we exclude this setting in Figure~\ref{fig:scaling}.

\begin{table}[!t]
\caption{Ablation study of modules in DeepGate3}
    \vspace{-5pt}
\label{tab:ablation}
\begin{tabular}{@{}l|l|l|l|l@{}}
\toprule
Setting                     & $L_{all}$ & $P^{tt}$ & $P^{con}$ & $P^{in}$ \\ \midrule
DeepGate3                   & \textbf{0.8553 }      & 0.0237 & \textbf{93.32\%}    & \textbf{95.18\%}         \\
DG3 w/o RT             & 1.3054       & \textbf{0.0226} & 85.76\%    & 82.39\%         \\
DG3 w/o PT            & 1.5033            &0.0731        &92.04\%           &88.14\%    \\
DG3 w/o RT \& PT & 2.2215       & 0.0898 & 85.58\%    & 78.57\%         \\ \bottomrule
\end{tabular}
\end{table}

We conclude three observations from the Table~\ref{tab:scaling}. First, DeepGate2 shows limited scalability when facing large dataset. 
In Table~\ref{tab:scaling}, for DeepGate3, most of the pre-training tasks achieve their best performance with $100\%$ of the data. In contrast, for DeepGate2, optimal performance across these tasks occurs at $5\%$, $30\%$, $50\%$, and $100\%$.
Furthermore, as illustrated in Figure~\ref{fig:scaling},the overall loss and \(P^{tt}\) for DeepGate2 no longer decrease when the data fraction increases from $50\%$ to $100\%$. However, for DeepGate3, both the overall loss and \(P^{tt}\) continue to decrease. In terms of \(P^{con}\) and \(P^{in}\), while DeepGate2 shows minimal improvement with additional data, DeepGate3 consistently enhances performance when trained with more data. 

Second, the loss reduction rate of DeepGate3 is also superior to that of DeepGate2. As the data increases from $5\%$ to $100\%$, these two methods exhibit different growth rates. For DeepGate2, the overall loss reduction ratios between $5\%$ data and $100\%$ data are $15.06\%$, $8.45\%$, $10.56\%$, and $-3.37\%$; for DeepGate3, the reduction ratios are $13.67\%$, $27.23\%$, $10.67\%$, and $20.02\%$.  

Last, comparing the best performances of each model, i.e., DeepGate2 with $50\%$ data and DeepGate3 with $100\%$ data, DeepGate3 consistently demonstrates superior performance. For example, considering the evaluation metrics in Section~\ref{sec:eval_metric}, the improvements in \(L_{all}\), \(P^{tt}\), \(P^{con}\), and \(P^{in}\) for DeepGate3 are 1.2938, 0.0661, 7.74\%, and 16.61\% respectively, showcasing the promising capabilities of DeepGate3.

\subsection{Ablation Study}
In this section, we perform ablation studies on the primary components of DeepGate3, namely the Refine Transformer (RT) and the Pooling Transformer (PT), following the metrics outlined in Section~\ref{sec:eval_metric}.
For DeepGate3 without the RT (\textit{DG3 w/o RT}), we use the embedding of DeepGate2 and PT to perform graph-level pre-training tasks. For DeepGate3 without the PT (\textit{DG3 w/o PT}), we replace PT with average pooling. For DeepGate3 without both RT and PT (\textit{DG3 w/o RT \& PT}), we only use the embedding of DeepGate2 and average pooling.


\noindent \textbf{Effectiveness of Refine Transformer. }
We evaluated the performance difference between DeepGate3 and DeepGate2, focusing on the inclusion of the PT. As demonstrated in Table~\ref{tab:ablation}, when comparing to \textit{DG3 w/o RT}, DeepGate3 shows enhancements in \(P^{con}\) and \(P^{in}\) by 7.56\% and 12.79\% respectively, and a reduction in overall loss by 0.4501, while maintaining comparable performance in \(P^{tt}\). Similarly, \textit{DG3 w/o PT} outperforms \textit{DG3 w/o RT \& PT} in terms of \(L_{all}\), \(P^{tt}\), \(P^{con}\), and \(P^{in}\), with improvements of 0.7182, 0.167, 6.46\%, and 9.57\% respectively. These results highlight the critical role of the Refine Transformer in refining initial embeddings and capturing complex, long-term correlations between gates.

\noindent \textbf{Effectiveness of Pooling function. } 
Prior work on graph-level tasks primarily utilized average or max pooling, such as Gamora~\cite{wu2023gamora} and HOGA~\cite{deng2024less}, which often fails to capture circuit-specific information and struggles to perform well under varying conditions. Our proposed Pooling Transformer addresses this by incorporating special tokens to account for diverse circumstances. According to the results in Table~\ref{tab:ablation}, the inclusion of the Pooling Transformer significantly enhances \(L_{all}\), \(P^{tt}\), and \(P^{in}\) in DeepGate3 by 0.6480, 0.0494, and 7.04\% respectively. Note that the Pooling Transformer is primarily related to graph-level tasks. Consequently, DeepGate3 and \textit{DG3 w/o PT} exhibit similar performance on gate-level tasks, such as \(P^{con}\).
Moreover, when comparing the performance of \textit{DG3 w/o RT} to that of \textit{DG3 w/o RT \& PT}, we observe better performance in \(P^{tt}\) and \(P^{in}\) by 0.0672 and 3.82\%, respectively. Since \(P^{con}\) is a gate-level task unrelated to the Pooling Transformer, the performance between \textit{DG3 w/o RT} and \textit{DG3 w/o RT \& PT} remains almost unchanged.

\subsection{Downstream Task: SAT Solving}

The Boolean Satisfiability (SAT) problem is a fundamental problem that determines whether a Boolean formula can output logic-1 with at least one variable assignment. It is the first proven NP-complete problem~\cite{cook1971complexity} and serves as a foundational problem applicable to various fields, including scheduling, planning and verification. The prevalent SAT solvers opt for the conflict-driven clause learning (CDCL) as the backbone solving algorithm. CDCL efficiently handles searching path conflicts and effectively explores additional constraints to reduce searching space. Over the past decades, numerous heuristic designs~\cite{lu2003circuit, audemard2018glucose, kurin2020can, shi2023satformer} have been proposed to accelerate CDCL process in SAT solvers. 

\noindent\textbf{Solving heuristic Design.} ~\cite{lu2003circuit} introduces a variable decision heuristic that assigns the opposite value to a pair of correlated variables, aiming to intentionally cause conflicts rapidly. While this approach achieves notable speed improvements compared to traditional SAT solvers, it still relies on time-consuming logic simulation to acquire the necessary functional relation between variables. DeepGate2~\cite{shi2023deepgate2} proposes an alternative solution by utilizing functional embeddings to measure the similarity between variables based on their functionality. Similarly, we can leverage the gate-level embeddings to predict functional similarity and identify correlated variables. 

\begin{figure} [!t]
    \centering
    \includegraphics[width=0.8\linewidth]{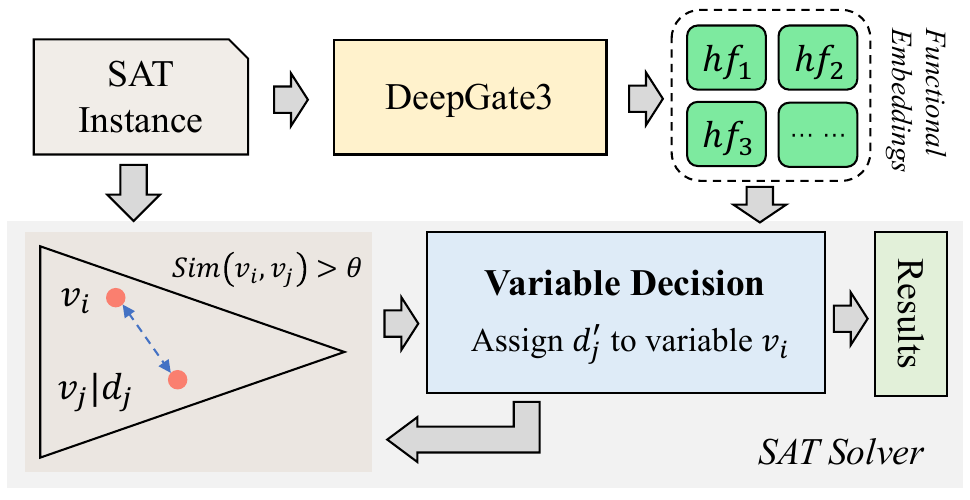}
    \vspace{-5pt}
    \caption{The pipeline of variable decision heuristics for SAT solving}
    \vspace{-10pt}
    \label{fig:SAT}
\end{figure}

The variable decision heuristic pipeline is shown in Figure~\ref{fig:SAT}. Given a SAT instance, the first step is to obtain gate-level functional embeddings prior to solving. During the variable decision process, a decision value $d_i$ is assigned to variable $v_i$. If another variable $v_j$ with an assigned value $d_j$ is identified as correlated to $v_i$, the reversed value $d_j'$ is assigned to $v_i$, i.e., $d_i = 0\ if\ d_j = 1$ or $d_i = 1\ if\ d_j = 0$. The determination of correlated variables relies on their functional similarity, as defined in Eq.~\eqref{Eq:funcSim}, where the similarity $Sim(v_i, v_j)$ exceeding the threshold $\theta$ indicates correlation.
\begin{equation} \label{Eq:funcSim}
    Sim(v_i, v_j) = MLP_{gate\_tt}(hf_{v_i}, hf_{v_j})
\end{equation}

\noindent\textbf{Experiment Settings. } We utilize the CaDiCal~\cite{queue2019cadical} SAT solver as the backbone solver and modify the variable decision heuristic based on it. In the Baseline setting, SAT problems are directly solved using the backbone SAT solver. In the DeepGate3 setting, our model is further fine-tuned efficiently on large-scale circuits in our proposed shifting-window manner (see Section~\ref{Sec:Method:Scale}). We fine-tune DeepGate3 for $200$ epochs only with the pair-wise truth table distance prediction task, i.e., optimizing the loss function $L_{gate}^{tt\_pair}$ as Eq.~\eqref{Eq:loss:gatefunc}. Then, we enhance the variable decision heuristic by incorporating gate-level embeddings produced by fine-tuned DeepGate3. Additionally, we also employ the embeddings obtained by DeepGate2 for comparison. The test cases, denoted as C1-C5, are collected from industrial logic equivalence checking (LEC) problems. These cases exhibit diverse levels of solving complexity, showcasing a range of challenges encountered in practical scenarios. 

\noindent\textbf{Results. } Table~\ref{TAB:SAT} presents a runtime comparison among the Baseline, DeepGate2, and DeepGate3 settings, where the runtime reduction compared to the Baseline setting is denoted as Red. and the number of gates in each circuit case is represented as \# Gates. Based on the observations from the table, we can draw three main conclusions. First, both DeepGate2 and DeepGate3 capture the functional correlation of variables to guide variable decision, resulting in reduced solving time compared to the Baseline setting. On average, DeepGate2 achieves a runtime reduction of $14.92\%$, and DeepGate3 achieves average reduction of $50.46\%$. However, it is worth noting that the heuristic-based solving strategy is less effective for easier cases, as the model inference process accounts for a significant portion of the total runtime. Secondly, DeepGate3 demonstrates its capability to handle large-scale circuits, as the LEC test cases contain more than $10$K gates, significantly surpassing the size of circuits in the pre-training dataset (with an average of $225.29$ gates). Thirdly, DeepGate3 exhibits superior performance compared to DeepGate2 in this task, indicating that DeepGate3 captures more informative gate-level functional embeddings. In conclusion, our model demonstrates effective generalization ability to solve practical SAT solving problems. 

\begin{table}[!t]
\caption{The comparison of SAT solving runtime} \label{TAB:SAT}
\vspace{-5pt}
\setlength\tabcolsep{3.0pt}
\begin{tabular}{@{}ll|l|ll|ll@{}}
\toprule
              &           & Baseline        & \multicolumn{2}{c|}{DeepGate2}     & \multicolumn{2}{c}{DeepGate3}      \\
Case          & \# Gates  & Time (s)        & Time (s)        & Red.             & Time (s)        & Red.             \\ \midrule
C1            & 10,921    & 15.93           & 15.18           & 4.71\%           & 16.74           & -5.08\%          \\
C2            & 13,955    & 38.56           & 34.34           & 10.94\%          & 20.20           & 47.61\%          \\
C4            & 19,369    & 108.53          & 94.59           & 12.84\%          & 105.60          & 2.70\%           \\
C5            & 14,496    & 574.03          & 425.72          & 25.84\%          & 406.46          & 29.19\%          \\
C6            & 19,469    & 1960.91         & 1725.72         & 11.99\%          & 787.64          & 59.83\%          \\ \midrule
\textbf{Avg.} & \textbf{} & \textbf{539.59} & \textbf{459.11} & \textbf{14.92\%} & \textbf{267.33} & \textbf{50.46\%} \\ \bottomrule
\end{tabular}
\vspace{-10pt}
\end{table}

\section{Conclusion}
\label{Sec:Conclusion}
In this work, we introduced DeepGate3, a pioneering framework that leverages the synergy between GNNs and Transformers to address the scalability of circuit representation learning. DeepGate3's innovative architecture, featuring Refine Transformer and Pooling Transformer mechanisms, significantly enhances the scalability and generalization capabilities of circuit representation learning. Moreover, the introduction of multiple novel supervision tasks has enriched the learning process, allowing DeepGate3 to capture a broader range of circuit behaviors with higher fidelity. Our experimental results demonstrate that DeepGate3 not only outperforms its predecessors DeepGate2 but also sets new benchmarks in handling complex and large-scale circuit designs efficiently. Future work will focus on incorporating additional data types, such as temporal and operational conditions, to enrich model insights and expanding DeepGate3's applications within EDA tasks.

\balance

\section*{Acknowledgments}
This work was supported in part by the General Research Fund of the Hong Kong Research Grants Council (RGC) under Grant No. 14212422. 

\bibliographystyle{unsrt}
\bibliography{reference}

\begin{thebibliography}{10}

\bibitem{chowdhery2023palm}
Aakanksha Chowdhery, Sharan Narang, Jacob Devlin, Maarten Bosma, Gaurav Mishra, Adam Roberts, Paul Barham, Hyung~Won Chung, Charles Sutton, Sebastian Gehrmann, et~al.
\newblock Palm: Scaling language modeling with pathways.
\newblock {\em Journal of Machine Learning Research}, 24(240):1--113, 2023.

\bibitem{achiam2023gpt}
Josh Achiam, Steven Adler, Sandhini Agarwal, Lama Ahmad, Ilge Akkaya, Florencia~Leoni Aleman, Diogo Almeida, Janko Altenschmidt, Sam Altman, Shyamal Anadkat, et~al.
\newblock Gpt-4 technical report.
\newblock {\em arXiv preprint arXiv:2303.08774}, 2023.

\bibitem{t5}
Colin Raffel, Noam Shazeer, Adam Roberts, Katherine Lee, Sharan Narang, Michael Matena, Yanqi Zhou, Wei Li, and Peter~J Liu.
\newblock Exploring the limits of transfer learning with a unified text-to-text transformer.
\newblock {\em Journal of machine learning research}, 21(140):1--67, 2020.

\bibitem{liu2019roberta}
Yinhan Liu, Myle Ott, Naman Goyal, Jingfei Du, Mandar Joshi, Danqi Chen, Omer Levy, Mike Lewis, Luke Zettlemoyer, and Veselin Stoyanov.
\newblock Roberta: A robustly optimized bert pretraining approach.
\newblock {\em arXiv preprint arXiv:1907.11692}, 2019.

\bibitem{wang2023internimage}
Wenhai Wang, Jifeng Dai, Zhe Chen, Zhenhang Huang, Zhiqi Li, Xizhou Zhu, Xiaowei Hu, Tong Lu, Lewei Lu, Hongsheng Li, et~al.
\newblock Internimage: Exploring large-scale vision foundation models with deformable convolutions.
\newblock In {\em Proceedings of the IEEE/CVF Conference on Computer Vision and Pattern Recognition}, pages 14408--14419, 2023.

\bibitem{liu2023towards}
Jiawei Liu, Cheng Yang, Zhiyuan Lu, Junze Chen, Yibo Li, Mengmei Zhang, Ting Bai, Yuan Fang, Lichao Sun, Philip~S Yu, et~al.
\newblock Towards graph foundation models: A survey and beyond.
\newblock {\em arXiv preprint arXiv:2310.11829}, 2023.

\bibitem{li2022deepgate}
Min Li, Sadaf Khan, Zhengyuan Shi, Naixing Wang, Huang Yu, and Qiang Xu.
\newblock Deepgate: Learning neural representations of logic gates.
\newblock In {\em Proceedings of the 59th ACM/IEEE Design Automation Conference}, pages 667--672, 2022.

\bibitem{shi2023deepgate2}
Zhengyuan Shi, Hongyang Pan, Sadaf Khan, Min Li, Yi~Liu, Junhua Huang, Hui-Ling Zhen, Mingxuan Yuan, Zhufei Chu, and Qiang Xu.
\newblock Deepgate2: Functionality-aware circuit representation learning.
\newblock In {\em 2023 IEEE/ACM International Conference on Computer Aided Design (ICCAD)}, pages 1--9. IEEE, 2023.

\bibitem{khan2023deepseq}
Sadaf Khan, Zhengyuan Shi, Min Li, and Qiang Xu.
\newblock Deepseq: Deep sequential circuit learning.
\newblock {\em arXiv preprint arXiv:2302.13608}, 2023.

\bibitem{wu2023gamora}
Nan Wu, Yingjie Li, Cong Hao, Steve Dai, Cunxi Yu, and Yuan Xie.
\newblock Gamora: Graph learning based symbolic reasoning for large-scale boolean networks.
\newblock In {\em 2023 60th ACM/IEEE Design Automation Conference (DAC)}, pages 1--6. IEEE, 2023.

\bibitem{deng2024less}
Chenhui Deng, Zichao Yue, Cunxi Yu, Gokce Sarar, Ryan Carey, Rajeev Jain, and Zhiru Zhang.
\newblock Less is more: Hop-wise graph attention for scalable and generalizable learning on circuits.
\newblock {\em arXiv preprint arXiv:2403.01317}, 2024.

\bibitem{kaplan2020scaling}
Jared Kaplan, Sam McCandlish, Tom Henighan, Tom~B Brown, Benjamin Chess, Rewon Child, Scott Gray, Alec Radford, Jeffrey Wu, and Dario Amodei.
\newblock Scaling laws for neural language models.
\newblock {\em arXiv preprint arXiv:2001.08361}, 2020.

\bibitem{liu2024neural}
Jingzhe Liu, Haitao Mao, Zhikai Chen, Tong Zhao, Neil Shah, and Jiliang Tang.
\newblock Neural scaling laws on graphs.
\newblock {\em arXiv preprint arXiv:2402.02054}, 2024.

\bibitem{alon2020bottleneck}
Uri Alon and Eran Yahav.
\newblock On the bottleneck of graph neural networks and its practical implications.
\newblock {\em arXiv preprint arXiv:2006.05205}, 2020.

\bibitem{topping2021understanding}
Jake Topping, Francesco Di~Giovanni, Benjamin~Paul Chamberlain, Xiaowen Dong, and Michael~M Bronstein.
\newblock Understanding over-squashing and bottlenecks on graphs via curvature.
\newblock {\em arXiv preprint arXiv:2111.14522}, 2021.

\bibitem{kipf2016semi}
Thomas~N Kipf and Max Welling.
\newblock Semi-supervised classification with graph convolutional networks.
\newblock {\em arXiv preprint arXiv:1609.02907}, 2016.

\bibitem{hamilton2017inductive}
Will Hamilton, Zhitao Ying, and Jure Leskovec.
\newblock Inductive representation learning on large graphs.
\newblock {\em Advances in neural information processing systems}, 30, 2017.

\bibitem{xu2018powerful}
Keyulu Xu, Weihua Hu, Jure Leskovec, and Stefanie Jegelka.
\newblock How powerful are graph neural networks?
\newblock {\em arXiv preprint arXiv:1810.00826}, 2018.

\bibitem{vaswani2017attention}
Ashish Vaswani, Noam Shazeer, Niki Parmar, Jakob Uszkoreit, Llion Jones, Aidan~N Gomez, {\L}ukasz Kaiser, and Illia Polosukhin.
\newblock Attention is all you need.
\newblock {\em Advances in neural information processing systems}, 30, 2017.

\bibitem{yun2019transformers}
Chulhee Yun, Srinadh Bhojanapalli, Ankit~Singh Rawat, Sashank~J Reddi, and Sanjiv Kumar.
\newblock Are transformers universal approximators of sequence-to-sequence functions?
\newblock {\em arXiv preprint arXiv:1912.10077}, 2019.

\bibitem{zhang2020graph}
Jiawei Zhang, Haopeng Zhang, Congying Xia, and Li~Sun.
\newblock Graph-bert: Only attention is needed for learning graph representations.
\newblock {\em arXiv preprint arXiv:2001.05140}, 2020.

\bibitem{ying2021transformers}
Chengxuan Ying, Tianle Cai, Shengjie Luo, Shuxin Zheng, Guolin Ke, Di~He, Yanming Shen, and Tie-Yan Liu.
\newblock Do transformers really perform badly for graph representation?
\newblock {\em Advances in neural information processing systems}, 34:28877--28888, 2021.

\bibitem{rong2020self}
Yu~Rong, Yatao Bian, Tingyang Xu, Weiyang Xie, Ying Wei, Wenbing Huang, and Junzhou Huang.
\newblock Self-supervised graph transformer on large-scale molecular data.
\newblock {\em Advances in neural information processing systems}, 33:12559--12571, 2020.

\bibitem{xia2024opengraph}
Lianghao Xia, Ben Kao, and Chao Huang.
\newblock Opengraph: Towards open graph foundation models.
\newblock {\em arXiv preprint arXiv:2403.01121}, 2024.

\bibitem{shi2022deeptpi}
Zhengyuan Shi, Min Li, Sadaf Khan, Liuzheng Wang, Naixing Wang, Yu~Huang, and Qiang Xu.
\newblock Deeptpi: Test point insertion with deep reinforcement learning.
\newblock In {\em 2022 IEEE International Test Conference (ITC)}, pages 194--203. IEEE, 2022.

\bibitem{li2023eda}
Min Li, Zhengyuan Shi, Qiuxia Lai, Sadaf Khan, Shaowei Cai, and Qiang Xu.
\newblock On eda-driven learning for sat solving.
\newblock In {\em 2023 60th ACM/IEEE Design Automation Conference (DAC)}, pages 1--6. IEEE, 2023.

\bibitem{shi2024eda}
Zhengyuan Shi, Tiebing Tang, Sadaf Khan, Hui-Ling Zhen, Mingxuan Yuan, Zhufei Chu, and Qiang Xu.
\newblock Eda-driven preprocessing for sat solving.
\newblock {\em arXiv preprint arXiv:2403.19446}, 2024.

\bibitem{li2018deeper}
Qimai Li, Zhichao Han, and Xiao-Ming Wu.
\newblock Deeper insights into graph convolutional networks for semi-supervised learning.
\newblock In {\em Proceedings of the AAAI conference on artificial intelligence}, volume~32, 2018.

\bibitem{velivckovic2017graph}
Petar Veli{\v{c}}kovi{\'c}, Guillem Cucurull, Arantxa Casanova, Adriana Romero, Pietro Lio, and Yoshua Bengio.
\newblock Graph attention networks.
\newblock {\em arXiv preprint arXiv:1710.10903}, 2017.

\bibitem{dong2022pace}
Zehao Dong, Muhan Zhang, Fuhai Li, and Yixin Chen.
\newblock Pace: A parallelizable computation encoder for directed acyclic graphs.
\newblock In {\em International Conference on Machine Learning}, pages 5360--5377. PMLR, 2022.

\bibitem{mckay1981practical}
Brendan~D McKay et~al.
\newblock Practical graph isomorphism.
\newblock 1981.

\bibitem{devlin2018bert}
Jacob Devlin, Ming-Wei Chang, Kenton Lee, and Kristina Toutanova.
\newblock Bert: Pre-training of deep bidirectional transformers for language understanding.
\newblock {\em arXiv preprint arXiv:1810.04805}, 2018.

\bibitem{bunke1997relation}
Horst Bunke.
\newblock On a relation between graph edit distance and maximum common subgraph.
\newblock {\em Pattern recognition letters}, 18(8):689--694, 1997.

\bibitem{efficient_transformer_survey}
Yi~Tay, Mostafa Dehghani, Dara Bahri, and Donald Metzler.
\newblock Efficient transformers: A survey.
\newblock {\em ACM Computing Surveys}, 55(6):1--28, 2022.

\bibitem{efficient_transformer1}
Jiezhong Qiu, Hao Ma, Omer Levy, Scott Wen-tau Yih, Sinong Wang, and Jie Tang.
\newblock Blockwise self-attention for long document understanding.
\newblock {\em arXiv preprint arXiv:1911.02972}, 2019.

\bibitem{efficient_transformer2}
Rewon Child, Scott Gray, Alec Radford, and Ilya Sutskever.
\newblock Generating long sequences with sparse transformers.
\newblock {\em arXiv preprint arXiv:1904.10509}, 2019.

\bibitem{efficient_transformer3}
Iz~Beltagy, Matthew~E Peters, and Arman Cohan.
\newblock Longformer: The long-document transformer.
\newblock {\em arXiv preprint arXiv:2004.05150}, 2020.

\bibitem{efficient_transformer4}
Peter~J Liu, Mohammad Saleh, Etienne Pot, Ben Goodrich, Ryan Sepassi, Lukasz Kaiser, and Noam Shazeer.
\newblock Generating wikipedia by summarizing long sequences.
\newblock {\em arXiv preprint arXiv:1801.10198}, 2018.

\bibitem{efficient_transformer5}
Manzil Zaheer, Guru Guruganesh, Kumar~Avinava Dubey, Joshua Ainslie, Chris Alberti, Santiago Ontanon, Philip Pham, Anirudh Ravula, Qifan Wang, Li~Yang, et~al.
\newblock Big bird: Transformers for longer sequences.
\newblock {\em Advances in neural information processing systems}, 33:17283--17297, 2020.

\bibitem{ITC99}
Scott Davidson.
\newblock Characteristics of the itc’99 benchmark circuits.
\newblock In {\em ITSW}, 1999.

\bibitem{albrecht2005iwls}
Christoph Albrecht.
\newblock Iwls 2005 benchmarks.
\newblock In {\em IWLS}, 2005.

\bibitem{EPFLBenchmarks}
Luca Amar{\'u}, Pierre-Emmanuel Gaillardon, and Giovanni De~Micheli.
\newblock The epfl combinational benchmark suite.
\newblock In {\em IWLS}, number CONF, 2015.

\bibitem{takeda2008opencore}
Opencores Team.
\newblock Opencores.
\newblock \url{https://opencores.org/}.

\bibitem{thakur2023benchmarking}
Shailja Thakur, Baleegh Ahmad, Zhenxing Fan, Hammond Pearce, Benjamin Tan, Ramesh Karri, Brendan Dolan-Gavitt, and Siddharth Garg.
\newblock Benchmarking large language models for automated verilog rtl code generation.
\newblock In {\em 2023 Design, Automation \& Test in Europe Conference \& Exhibition (DATE)}, pages 1--6. IEEE, 2023.

\bibitem{brayton2010abc}
Robert Brayton and Alan Mishchenko.
\newblock Abc: An academic industrial-strength verification tool.
\newblock In {\em Computer Aided Verification: 22nd International Conference, CAV 2010, Edinburgh, UK, July 15-19, 2010. Proceedings 22}, pages 24--40. Springer, 2010.

\bibitem{scaling1}
Xiaohua Zhai, Alexander Kolesnikov, Neil Houlsby, and Lucas Beyer.
\newblock Scaling vision transformers.
\newblock In {\em Proceedings of the IEEE/CVF conference on computer vision and pattern recognition}, pages 12104--12113, 2022.

\bibitem{scaling2}
Yamini Bansal, Behrooz Ghorbani, Ankush Garg, Biao Zhang, Colin Cherry, Behnam Neyshabur, and Orhan Firat.
\newblock Data scaling laws in nmt: The effect of noise and architecture.
\newblock In {\em International Conference on Machine Learning}, pages 1466--1482. PMLR, 2022.

\bibitem{scaling3}
Niklas Muennighoff, Alexander Rush, Boaz Barak, Teven Le~Scao, Nouamane Tazi, Aleksandra Piktus, Sampo Pyysalo, Thomas Wolf, and Colin~A Raffel.
\newblock Scaling data-constrained language models.
\newblock {\em Advances in Neural Information Processing Systems}, 36, 2024.

\bibitem{scaling4}
Ibrahim~M Alabdulmohsin, Behnam Neyshabur, and Xiaohua Zhai.
\newblock Revisiting neural scaling laws in language and vision.
\newblock {\em Advances in Neural Information Processing Systems}, 35:22300--22312, 2022.

\bibitem{cook1971complexity}
Stephen~A Cook.
\newblock The complexity of theorem-proving procedures.
\newblock In {\em Proceedings of the third annual ACM symposium on Theory of computing}, pages 151--158, 1971.

\bibitem{lu2003circuit}
Feng Lu, L-C Wang, Kwang-Ting Cheng, and RC-Y Huang.
\newblock A circuit sat solver with signal correlation guided learning.
\newblock In {\em 2003 Design, Automation and Test in Europe Conference and Exhibition}, pages 892--897. IEEE, 2003.

\bibitem{audemard2018glucose}
Gilles Audemard and Laurent Simon.
\newblock On the glucose sat solver.
\newblock {\em International Journal on Artificial Intelligence Tools}, 27(01):1840001, 2018.

\bibitem{kurin2020can}
Vitaly Kurin, Saad Godil, Shimon Whiteson, and Bryan Catanzaro.
\newblock Can q-learning with graph networks learn a generalizable branching heuristic for a sat solver?
\newblock {\em Advances in Neural Information Processing Systems}, 33:9608--9621, 2020.

\bibitem{shi2023satformer}
Zhengyuan Shi, Min Li, Yi~Liu, Sadaf Khan, Junhua Huang, Hui-Ling Zhen, Mingxuan Yuan, and Qiang Xu.
\newblock Satformer: Transformer-based unsat core learning.
\newblock In {\em 2023 IEEE/ACM International Conference on Computer Aided Design (ICCAD)}, pages 1--4. IEEE, 2023.

\bibitem{queue2019cadical}
SEPARATE~DECISION QUEUE.
\newblock Cadical at the sat race 2019.
\newblock {\em SAT RACE 2019}, page~8, 2019.

\end{thebibliography}

\end{document}